# Leveraging Automated Machine Learning for Text Classification: Evaluation of AutoML Tools and Comparison with Human Performance

Matthias Blohm[1][a], Marc Hanussek[1][b] and Maximilien Kintz[2][c]
[1]*University of Stuttgart, Institute of Human Factors and Technology Management (IAT), Stuttgart, Germany*
[2]*Fraunhofer IAO, Fraunhofer Institute for Industrial Engineering IAO, Stuttgart, Germany*
*{matthias.blohm, marc.hanussek}@iat.uni-stuttgart.de, maximilien.kintz@iao.fraunhofer.de*

Abstract: Recently, Automated Machine Learning (AutoML) has registered increasing success with respect to tabular data. However, the question arises whether AutoML can also be applied effectively to text classification tasks. This work compares four AutoML tools on 13 different popular datasets, including Kaggle competitions, and opposes human performance. The results show that the AutoML tools perform better than the machine learning community in 4 out of 13 tasks and that two stand out.

## 1 INTRODUCTION

With recent progress in Automated Machine Learning (AutoML) technologies, the question arises whether current systems and tools can beat state-of-the-art results achieved by human data scientists.

While a lot of work has been seen for benchmarking structured resp. tabular datasets (He et al., 2019; Truong et al., 2019; Zöller & Huber, 2019), the application of AutoML for natural language processing (NLP) tasks like text classification has not gained that much attention yet.

This is underlined by the fact that as of now, many popular open source AutoML libraries do not provide any support for processing raw text input samples. Instead, text input needs to be converted to structured data manually, for instance as word or sentence embeddings, before feeding them into AutoML libraries.

Nonetheless, many operators, including enterprises, are interested in building AI-based text or document classification solutions, e.g. in the area of incoming daily post in form of emails or letters that are predestined for automated tasks of pre-categorization. Since the realization of such solutions usually requires deep knowledge about appropriate text pre-processing and model building techniques, which often are not present, the use of AutoML tools might be a good (first) approach for many use cases.

In our work we aim to evaluate the current performance of four popular AutoML tools on the task of text classification for 13 common English textual datasets and competitions. On the one hand, we compare performance between tools and datasets. On the other hand, we give insights about AutoML performance in general against best known scores achieved by human data scientists using classical machine learning.

The paper is structured as follows: In Section 2 we list related work in the field of AutoML applied for NLP tasks. In Section 3, methodoglogy and settings for our experiments are described. Discussion of our results and analysis are given in Section 4, followed by a conclusion and description of intended future work in Section 5.

## 2 RELATED WORK

AutoML services optimized for NLP tasks like Amazon Comprehend (Mishra) already exist on the market. While these products are specialized in solving problems of text classification or named entity recognition, popular open source tools like auto-sklearn still lack those capabilities and focus on tabular data only.
However, some benchmarks have been published comparing performance of different AutoML approaches on multiple tasks and datasets, of which some are also textual and which provide good first insights about the current state-of-the-art in the domain of AutoML (He et al., 2019).

[a] https://orcid.org/0000-0000-0000-0000
[b] https://orcid.org/0000-0002-8041-8858
[c] https://orcid.org/0000-0000-0000-0000

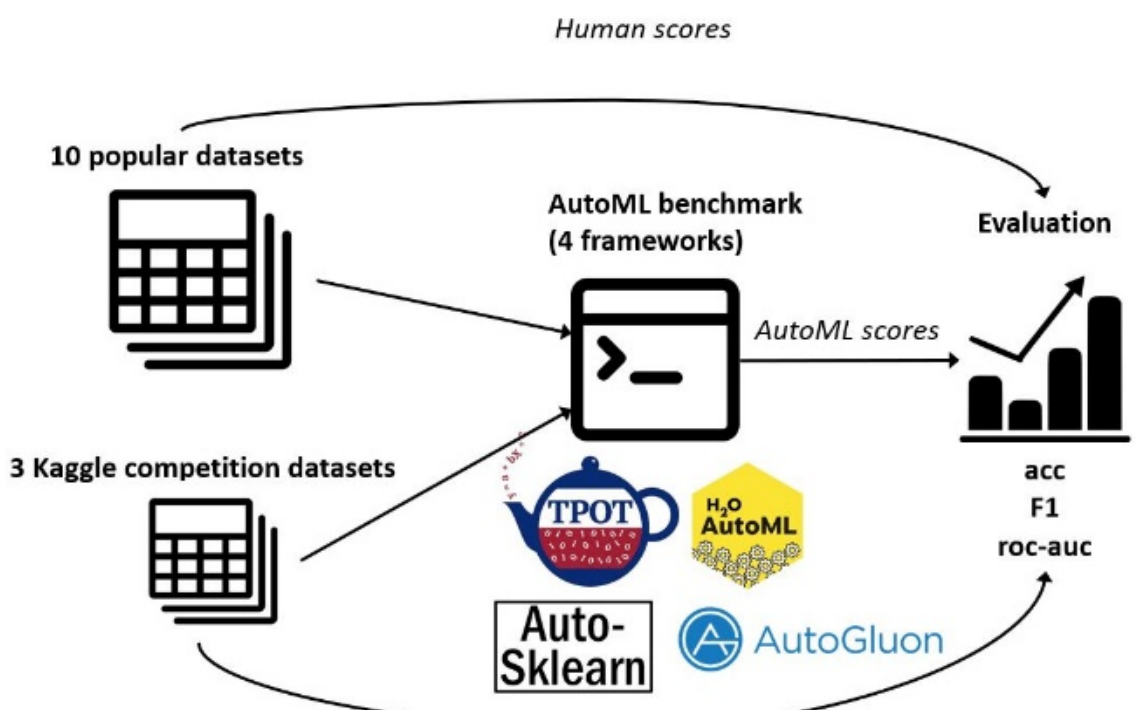

Figure 1: Schematic depiction of our approach

Additionally, some individual works already tackled the task of automated representation and processing of text contents (Madrid et al., 2019). While Wong et al. (2018) explored the usage of transfer learning for AutoML text classification tasks, Estevez-Velarde et al. (2019) experimented with grammatical evolution strategies to extract knowledge from Spanish texts.

Furthermore, Drori et al. (2019) combined language embeddings created from metadata files with AutoML tools to improve result performance.

# 3 METHODOLOGY OF THE BENCHMARK

In this section we describe our approach and experiment settings for the evaluation of AutoML tools on text classification datasets. Figure 1 illustrates this process. In a first step, we collected suitable datasets together with available human performances on this task. In a second step and after data pre-processing, we used automated machine learning to let each tool find the best model for the task. In a final step, we evaluated tool performances and compared overall AutoML scores to the best-known results achieved without usage of AutoML (human performance).

## 3.1 Datasets

We considered 13 publicly available datasets of which three were used in past Kaggle competitions. The datasets cover a wide range of topics including sentiment analysis, fake news or fake posts detection and categorization of everyday as well as scientific texts. The language is English except for the *Roman Urdu dataset*, which displays the Urdu language written with the Roman script. The number of samples in each dataset ranges from 804 (Math lectures) to 1600000 (*Sentiment140 dataset* with 1.6 million tweets) with the average being 428295. The length of the texts differs significantly within individual datasets as well as across datasets: The dataset with the greatest range between shortest and longest text is *20_newsgroup* with a minimum of 1 character and a maximum of 156224 characters. On average, the shortest texts can be found in *Real or Not? NLP with Disaster Tweets* (195 characters), the longest texts are again located in *20_newsgroup* (156224 characters). The number of target classes varies from two (binary classification) to 20. The average number of classes is six.

## 3.2 Data Preparation

All datasets were pre-processed in such a way that only two columns remained: text and target column. In order to do so, we removed all other data such as IDs or other meta data and merged different text columns, if suitable.

As of now, most of the AutoML libraries have no native support for raw text input processing. Therefore, we decided to use the transformer tool provided by Reimers and Gurevych (2019) to represent all datasets as structured BERT embeddings using a generic English model without any fine tuning. In detail, every text sample was encoded as an embedding of size 768, allowing usage and comparability also among the tools without support for unstructured text inputs.

Nevertheless, we are aware that the choice of a fixed textual representation model might have great influence on the quality of the resulting models. Hence, more fine-grained experiments with different embeddings or pre-processing steps remain an open point for our future work.

## 3.3 AutoML Benchmark

We obtained most of our results by using AutoML Benchmark v0.9 (Gijsbers et al., 2019). AutoML Benchmark is an open and extensible benchmark

Table 1: Best performing AutoML tools with and without AutoGluon Text

| Number | Best AutoML tool | Best AutoML tool except AutoGluon Text |
|---|---|---|
| 1 | auto-sklearn | auto-sklearn |
| 2 | AutoGluon Text | auto-sklearn |
| 3 | AutoGluon Text | auto-sklearn |
| 4 | AutoGluon Text | auto-sklearn |
| 5 | AutoGluon Text | auto-sklearn |
| 6 | auto-sklearn | auto-sklearn |
| 7 | AutoGluon Text | auto-sklearn |
| 8 | H2O | H2O |
| 9 | AutoGluon Text, TPOT | TPOT |
| 10 | auto-sklearn | auto-sklearn |
| 11 | AutoGluon Text | H2O |
| 12 | auto-sklearn | auto-sklearn |
| 13 | auto-sklearn | auto-sklearn |

framework which allows for comparing AutoML systems in a uniform way.

We ran the benchmarks with default settings as defined in config.yaml in the AutoML Benchmark project, i.e. usage of all cores, 2GiB of memory left to the OS, amount of memory computed from os available memory and many more. The only parameter we set was the runtime per fold, which we set to one hour.

## 3.4 AutoML Tools

For our experiments we evaluated the performance of four AutoML tools: TPOT v0.11.5 (Olson et al., 2016), H2O v3.30.0.4 (H2O.ai, 2017), auto-sklearn v0.5.2 (Feurer et al., 2015) and AutoGluon Text v0.0.14 (Erickson et al., 2020). We believe that this is a good mix of recent and older tools that partly come with support of deep learning technologies, too. For AutoGluon we used the built-in text prediction function (labelled as AutoGluon Text in this work) that allows raw text input, for the other libraries we pre-processed our data as described in Section 3.2. For reasons of comparison, we partly used AutoGluon Tabular as well. Generally, we consciously treated all tools as black boxes and without diving deeper into tool-specific algorithms and approaches. Therefore, we accept that AutoGluon Text might be at an advantage as is the only tool with support for raw text input.

## 3.5 Cross-Validation and Metrics

Table 3 lists the train-test-split configurations for each dataset as well as the primary metric that the models were optimized for and finally evaluated. For datasets having less than 50000 samples, we applied 5-fold cross validation and test size 25%, while for larger datasets we applied a train-test-split with test size 25%. Each split was created in a stratified fashion and using random shuffling of samples. For the cross-validation tasks, the final evaluation score was computed as the average over the results achieved by each of the 5 test splits.

## 3.6 Hardware

The machine we ran the benchmark on was a dedicated server we host locally. The server is equipped with two Intel Xeon Silver 4114 CPUs @2.20Ghz (yielding 20 cores in total), four 64GB DIMM DDR4 Synchronous 2666MHz memory modules and two NVIDIA GeForce GTX 1080 Ti (yielding more than 22GB VRAM in total).

# 4 RESULTS AND ANALYSIS

In this chapter, we state and discuss the main results.

**AutoML leaders**

The best performing AutoML tools are depicted in Table 1. The dataset ID number assignments are given in the data overview in Table 3. In seven out of 13 cases, AutoGluon Text performs best among AutoML tools, once coinciding with H2O. The second most successful tool is auto-sklearn with five first placements. TPOT and H2O lag far behind with one and zero wins, respectively. Note that AutoGluon Text did not complete one task while the others yielded results. When disregarding AutoGluon Text, as it operates on differently pre-processed data than the other four tools, auto-sklearn stands out with 10 out of 13 wins. H2O performs best twice, TPOT once.

Table 2: Best performing AutoML tools with and without AutoGluon Text

| Number | Best score by AutoML (acc) | Number of AutoML tools better or equal | Best known human score (acc) |
|---|---|---|---|
| 1 | 0.989 | 4 | 0.966[4] |
| 2 | 0.708 | 0 | 0.776[5] |
| 3 | 0.715 | 0 | 0.829[6] |
| 4 | 0.946 | 0 | 0.962[7] |
| 5 | 0.837 | 0 | 0.87[8] |
| 6 | 0.657 | 0 | 0.836[9] |
| 7 | 0.862 | 0 | 0.926[10] |
| 8 | 0.961 | 3 | 0.944[11] |
| 9 | 0.171 | 2 | 0.169[12] |
| 10 | 0.52 | 1 | 0.519[13] |
| 11 | 0.653 (F1) | 0 | 0.713[14] (F1) |
| 12 | 0.768 (F1) | 0 | 1[15] (F1) |
| 13 | 0.718 | 0 | 0.832[16] |

When we use the F1 score instead of the accuracy score, results differ in such a way that AutoGluon Text and auto-sklearn bring in five wins each, TPOT two and H2O one. Note that in three cases, AutoGluon Text was not able to calculate a F1 score.

**Human comparison**

In 9 out of 13 cases, the respective best AutoML tool cannot beat human performance. In particular, all three Kaggle competitions are won by humans. If AutoML outperforms humans, the average number of AutoML tools to do so is approximately 2.5.

**Quantitative analysis**

Considering aforementioned seven out of 13 cases, in which AutoGluon Text outperforms every other AutoML tool, its performance margin is averagely 8.7% relating to the respective runner-up. When undertaking this comparison on all 13 datasets, this margin is 2%. In the four out of 13 cases in which AutoML outperforms human performance, this margin accounts for 1.4% on average. This margin shrinks to -7.4% when considering all 13 datasets. The three Kaggle competitions stand out as, with 15.1%, the human advantage is considerably higher. In eight cases, we contrasted AutoGluon Text with AutoGluon Tabular. The Text Prediction feature performs better seven times and we observed the average margin to be 3.1%. Note that this margin varied considerably from -22.7% to 23.3%. An overview can be found in Table 2.

**Discussion**

It is understandable that the only AutoML tool featuring a text classification module (AutoGluon) wins this benchmark on the part of automated approaches. Regarding the cases in which AutoGluon Text outperforms the other AutoML tools, the fact that the margin is noteworthy emphasizes this circumstance. Thereby, the developers show that solutions specifically tailored to text classification indeed bring added value. On the contrary, auto-sklearn demonstrates that general-purpose tools do not necessarily lag too far behind. Since auto-sklearn is the only AutoML tool that can keep up with AutoGluon Text, one can understand this as relative strength of auto-sklearn or further potential of specifically tailored text classification modules.

When conducting machine learning benchmarks, one is confronted with the question which metrics to use. In this experiment, we made clear that there is no notable difference between accuracy and F1 score. Besides, we evaluated roc-auc-score and neither encountered major shifts concerning dominating AutoML tools. That suggests that our benchmark is not biased by the choice of evaluation metrics.

[4] https://www.kaggle.com/vennaa/notebook-spam-text-message-classification-with-r
[5] https://paperswithcode.com/sota/text-classification-on-yahoo-answers
[6] https://dl.acm.org/doi/fullHtml/10.1145/3329709
[7] https://www.kaggle.com/alexalex02/nlp-transformers-inference-optimization
[8] https://www.kaggle.com/menion/sentiment-analysis-with-bert-87-accuracy
[9] https://www.researchgate.net/publication/261040860_A_Novel_Semantic_Smoothing_Method_Based_on_Higher_Order_Paths_for_Text_Classification
[10] https://www.kaggle.com/arbazkhan971/sentiment-analysis-for-beginner-93-accuracy
[11] https://www.kaggle.com/kevinlwebb/cybertrolls-exploration-and-ml
[12] https://www.kaggle.com/sanskar27jain/kernelb49fc09b70
[13] https://www.kaggle.com/mohamadalhasan/fake-news-around-syrian-war
[14] https://www.kaggle.com/c/quora-insincere-questions-classification/leaderboard
[15] https://www.kaggle.com/c/nlp-getting-started/leaderboard
[16] https://www.kaggle.com/c/whats-cooking/leaderboard

The fact that all three Kaggle competitions are clearly won by humans is understandable since apparently, these are the tasks in which contestants put as much effort as possible. This circumstance was already observed in the work of (Hanussek et al., 2020) and it underlines the insight that, at this time, AutoML cannot beat humans in situation in which extraordinary results are required. This is again shown by our discovery, that concerning the cases in which AutoML outperforms humans, this outperformance is rather little and in most of the cases only one or two AutoML tools manage to do so (although the average is 2.5, which is attributable to the first task where all four AutoML tools beat human performance).

Finally, we want to address usage of the considered AutoML tools and AutoML benchmark. Occasionally, bold human intervention is required in order to make them work properly. Clearly, this is understandable as AutoML in general is a relatively new field and the tools are partly in early stage of development. However, it contradicts the idea of automated machine learning and we see great potential regarding stability, reliability and function range.

# 5. CONCLUSION AND OUTLOOK

The present works contributes to the standard of knowledge concerning AutoML performance in text classification. Our research interests were two-fold; comparison of performance between AutoML tools and confrontation with human performance. The results show that, in most cases, AutoML is not able to outperform humans in text classification. However, there are text classification tasks that can be solved better or equally by AutoML tools. With automated approaches becoming increasingly sophisticated, we see this disparity shrink in the future.

We see great potential in future development of specific text classification modules within AutoML tools. Such modules would further facilitate usage of machine learning by beginners and establishing a baseline for advanced users.

In the future, we will focus on investigating impact of different pre-processing techniques for texts (including more embedding types) for conclusive usage in AutoML tools. Evidently, there are more AutoML tools which should be evaluated, too.

Furthermore, testing AutoML for other NLP tasks like named entity recognition is an interesting topic for further research. Additionally, we will analyse performance of commercial cloud services that come with ready-to-use text classification functionality.

# REFERENCES

Almeida, T. A., Hidalgo, J. M. G., & Yamakami, A. (2011). Contributions to the Study of SMS Spam Filtering: New Collection and Results. In *DocEng '11, Proceedings of the 11th ACM Symposium on Document Engineering* (pp. 259–262). Association for Computing Machinery. https://doi.org/10.1145/2034691.2034742

Drori, I., Liu, L., Nian, Y., Koorathota, S., Li, J., Moretti, A., Freire, J., & Udell, M. (2019). *AutoML using Metadata Language Embeddings*.

Erickson, N., Mueller, J., Shirkov, A., Zhang, H., Larroy, P., Li, M., & Smola, A. (2020). AutoGluon-Tabular: Robust and Accurate AutoML for Structured Data. *ArXiv Preprint ArXiv:2003.06505*.

Estevez-Velarde, S., Gutiérrez, Y., Montoyo, A., & Almeida-Cruz, Y. (2019). AutoML Strategy Based on Grammatical Evolution: A Case Study about Knowledge Discovery from Text. In *Proceedings of the 57th Annual Meeting of the Association for Computational Linguistics* (pp. 4356–4365). Association for Computational Linguistics. https://doi.org/10.18653/v1/P19-1428

Feurer, M., Klein, A., Eggensperger, K., Springenberg, J., Blum, M., & Hutter, F. (2015). Efficient and Robust Automated Machine Learning. In C. Cortes, N. D. Lawrence, D. D. Lee, M. Sugiyama, & R. Garnett (Eds.), *Advances in Neural Information Processing Systems 28* (pp. 2962–2970). Curran Associates, Inc. http://papers.nips.cc/paper/5872-efficient-and-robust-automated-machine-learning.pdf

Gijsbers, P., LeDell, E., Poirier, S., Thomas, J., Bischl, B., & Vanschoren, J. (2019). An Open Source AutoML Benchmark. *ArXiv Preprint ArXiv:1907.00909 [Cs.LG]*.

Go, A., Bhayani, R., & Huang, L. (2009). Twitter Sentiment Classification using Distant Supervision. *Processing*, 1–6. http://www.stanford.edu/ alecmgo/papers/TwitterDistantSupervision09.pdf

H2O.ai. (2017). *H2O AutoML.* http://docs.h2o.ai/h2o/latest-stable/h2o-docs/automl.html

Hanussek, M., Blohm, M., & Kintz, M. (2020). *Can AutoML outperform humans? An evaluation on popular OpenML datasets using AutoML Benchmark.*

He, X., Zhao, K., & Chu, X. (2019, August 2). *AutoML: A Survey of the State-of-the-Art.* http://arxiv.org/pdf/1908.00709v4

Maas, A. L., Daly, R. E., Pham, P. T., Huang, D., Ng, A. Y., & Potts, C. (2011). Learning Word Vectors for Sentiment Analysis. In *Proceedings of the 49th Annual Meeting of the Association for Computational Linguistics: Human Language Technologies* (pp. 142–150). Association for Computational Linguistics. http://www.aclweb.org/anthology/P11-1015

Madrid, J., Escalante, H. J., & Morales, E. (2019). Meta-learning of textual representations. *CoRR, abs/1906.08934*.

Mehmood, K., Essam, D., Shafi, K., & Malik, M. K. (2019). Sentiment Analysis for a Resource Poor Language—Roman Urdu. *ACM Trans. Asian Low-Resour. Lang. Inf. Process.*, *19*(1). https://doi.org/10.1145/3329709

Mishra, A. Amazon Comprehend. In *Machine Learning in the AWS Cloud* (pp. 257–274). https://doi.org/10.1002/9781119556749.ch13

Olson, R. S., Bartley, N., Urbanowicz, R. J., & Moore, J. H. (2016). Evaluation of a Tree-based Pipeline Optimization Tool for Automating Data Science. In *GECCO '16, Proceedings of the Genetic and Evolutionary Computation Conference 2016* (pp. 485–492). ACM. https://doi.org/10.1145/2908812.2908918

Reimers, N., & Gurevych, I. (2019). Sentence-BERT: Sentence Embeddings using Siamese BERT-Networks. In *Proceedings of the 2019 Conference on Empirical Methods in Natural Language Processing.* Association for Computational Linguistics.

Truong, A., Walters, A., Goodsitt, J., Hines, K., Bruss, C. B., & Farivar, R. (2019). Towards Automated Machine Learning: Evaluation and Comparison of AutoML Approaches and Tools, 1471–1479. https://doi.org/10.1109/ICTAI.2019.00209

Wong, C., Houlsby, N., Lu, Y., & Gesmundo, A. (2018). Transfer Automatic Machine Learning. *CoRR, abs/1803.02780*.

Zöller, M.-A., & Huber, M. F. (2019). Survey on Automated Machine Learning. *CoRR, abs/1904.12054*.

# APPENDIX

Table 3: Datasets overview and statistics, CV = 5-fold cross validation

| **Number (ID)** | **Name** | **Source and/or Reference** | **CV** | **No. of classes** | **Primary Metric** |
|---|---|---|---|---|---|
| 1 | Spam Text Message Classification | https://www.kaggle.com/team-ai/spam-text-message-classification<br>(Almeida et al., 2011) | Yes | 2 | acc |
| 2 | Yahoo! Answers Topic Classification | https://github.com/LC-John/Yahoo-Answers-Topic-Classification-Dataset | No | 10 | acc |
| 3 | Roman Urdu Data Set Data Set | (Mehmood et al., 2019) | Yes | 2 | acc |
| 4 | Amazon Reviews for Sentiment Analysis | https://www.kaggle.com/bittlingmayer/amazonreviews | No | 2 | acc |
| 5 | Sentiment140 dataset with 1.6 million tweets | https://www.kaggle.com/kazanova/sentiment140<br>(Go et al., 2009) | No | 2 | acc |
| 6 | 20_newsgroup | https://scikit-learn.org/0.19/datasets/twenty_newsgroups.html | Yes | 20 | acc |
| 7 | IMDB Dataset of 50K Movie Reviews | https://www.kaggle.com/lakshmi25npathi/imdb-dataset-of-50k-movie-reviews<br>(Maas et al., 2011) | No | 2 | acc |
| 8 | Cyber Troll | https://zenodo.org/record/3665663 | Yes | 2 | acc |
| 9 | Math Lectures | https://www.kaggle.com/extralime/math-lectures | Yes | 11 | acc |
| 10 | FA-KES | https://www.kaggle.com/mohamadalhasan/a-fake-news-dataset-around-the-syrian-war | Yes | 2 | acc |
| 11 | Quora Insincere Questions Classification | https://www.kaggle.com/c/quora-insincere-questions-classification | No | 2 | acc |
| 12 | Real or Not? NLP with Disaster Tweets | https://www.kaggle.com/c/nlp-getting-started/data | Yes | 2 | F1 |
| 13 | What's Cooking? | https://www.kaggle.com/c/whats-cooking/overview | Yes | 20 | acc |